\documentclass[10pt,twocolumn,letterpaper]{article}

\usepackage[pagenumbers]{cvpr} 

\usepackage[dvipsnames]{xcolor}
\usepackage{multirow}
\usepackage{colortbl}

\usepackage{tcolorbox}
\newcommand{\code}[1]{\mbox{
    \ttfamily
    \tcbox[
        on line,
        boxsep=0pt, left=4pt, right=4pt, top=2pt, bottom=1.5pt,
        toprule=0pt, rightrule=0pt, bottomrule=0pt, leftrule=0pt,
        oversize=0pt, enlarge left by=0pt, enlarge right by=0pt,
        colframe=white, colback=black!12
    ]{#1}
}}

\newcommand*{\lineimg}[1]{%
    \raisebox{-.3\baselineskip}{%
        \includegraphics[
        height=\baselineskip,
        width=\baselineskip,
        keepaspectratio,
        ]{#1}%
    }%
}

\newcommand{\methodname}{\textsc{Video-ColBERT}\xspace}

\newcommand{\loss}{\mathcal{L}}
\newcommand{\bq}{\mathbf{q}}
\newcommand{\bv}{\mathbf{v}}
\newcommand{\bfr}{\mathbf{f}}

\usepackage{pifont}
\newcommand{\cmark}{\ding{51}}%
\newcommand{\xmark}{\ding{55}}

\usepackage{fancyhdr}

\definecolor{cvprblue}{rgb}{0.21,0.49,0.74}
\usepackage[pagebackref,breaklinks,colorlinks,allcolors=cvprblue]{hyperref}

\def\paperID{2079} 
\def\confName{CVPR}
\def\confYear{2025}

\title{Video-ColBERT: Contextualized Late Interaction for Text-to-Video Retrieval}

\author{Arun Reddy$^{1,2}$* \hspace{2.5mm} Alexander Martin$^{2}$* \hspace{2.5mm} Eugene Yang$^{2,3}$ \hspace{2.5mm} Andrew Yates$^{2,3}$ \hspace{2.5mm} Kate Sanders$^{2}$ \\
Kenton Murray$^{2,3}$ \hspace{2.5mm} Reno Kriz$^{2,3}$ \hspace{2.5mm} Celso M. de Melo$^{4}$ \hspace{2.5mm} Benjamin Van Durme$^{2,3}$ \hspace{2.5mm} Rama Chellappa$^{2}$ \\
\small ${}^1$Johns Hopkins Applied Physics Laboratory \hspace{2.5mm} ${}^2$Johns Hopkins University \hspace{2.5mm} \\ \small{${}^3$Human Language Technology Center of Excellence \hspace{2.5mm} ${}^4$DEVCOM Army Research Laboratory} \\
}

\begin{document}
\maketitle
\def\thefootnote{*}\footnotetext{Equal contribution.}
\def\thefootnote{\arabic{footnote}}
\begin{abstract}
In this work, we tackle the problem of text-to-video retrieval (T2VR). Inspired by the success of late interaction techniques in text-document, text-image, and text-video retrieval, our approach, Video-ColBERT, introduces a simple and efficient mechanism for fine-grained similarity assessment between queries and videos. Video-ColBERT is built upon three main components: a fine-grained spatial and temporal token-wise interaction, query and visual expansions, and a dual sigmoid loss during training. We find that this interaction and training paradigm leads to strong individual, yet compatible,  representations for encoding video content. These representations lead to increases in performance on common text-to-video retrieval benchmarks compared to other bi-encoder methods. 

\end{abstract}    
\fancypagestyle{firstpage}{
  \fancyhf{} 
  \renewcommand{\headrulewidth}{0pt} 
  \lfoot{} 
  \cfoot{\thepage} 
  \rfoot{\textit{Approved for public release. Distribution is unlimited.}} 
}

\thispagestyle{firstpage}

\section{Introduction}
\label{sec:intro}

With an ever-increasing amount of video data being generated and stored daily, the need for effective and efficient retrieval methods has become more pressing than ever. Text-to-video retrieval (T2VR) aims to address this by ranking large collections of videos based on their relevance to natural language queries. However, the task remains challenging due to the inherent modality gap between text and video representations. While recent advances in cross-modal retrieval have started to bridge this gap~\cite{chen2023VAST, li2024UnmaskedTeacher, wang2024InternVidDatsetViCLIPModel, wang2024InternVideo2, rizve2024VidLA}, significant progress is still needed to achieve reliable and scalable performance in real-world settings.

A common approach to efficient retrieval is the use of a bi-encoder method \cite{reimers2019sentence, karpukhin2020dense, radford2021CLIP}, where the query and document are encoded separately. 
Bi-encoders offer efficiency advantages over cross encoders \cite{lei2021ClipBERT, sun2019VideoBERT, gorti2022X-Pool, nogueira2020MonoT5}
because they do not require expensive interactions at retrieval time and can instead operate on a pre-computed index of the target data.

Some bi-encoder T2VR techniques \cite{luo2022CLIP4Clip, deng2023PromptSwitch} use single vectors to represent the text query and the video (\eg through mean pooling), then use a single dot product for similarity calculation at retrieval time. 

While this simple interaction may be sufficient for settings like language-image pre-training \cite{radford2021CLIP, zhai2022LiT, zhai2023SigLIP}, it can be challenging to encode video content and query concepts within single feature vectors \cite{sun2022LF-VILA, jin2022EMCL}. Other works \cite{wang2022DRL, ma2022X-CLIP} have applied more expressive interactions, like tokenwise interaction, to the video retrieval problem using CLIP \cite{radford2021CLIP} adapted for videos. We argue that these works are limited in their exploration of tokenwise interaction. Specifically, these approaches employ overly complicated interaction mechanisms or have limited final representation(s) used in their interactions. 
\begin{figure}[t]
  \centering
    \includegraphics[width=0.99\linewidth]{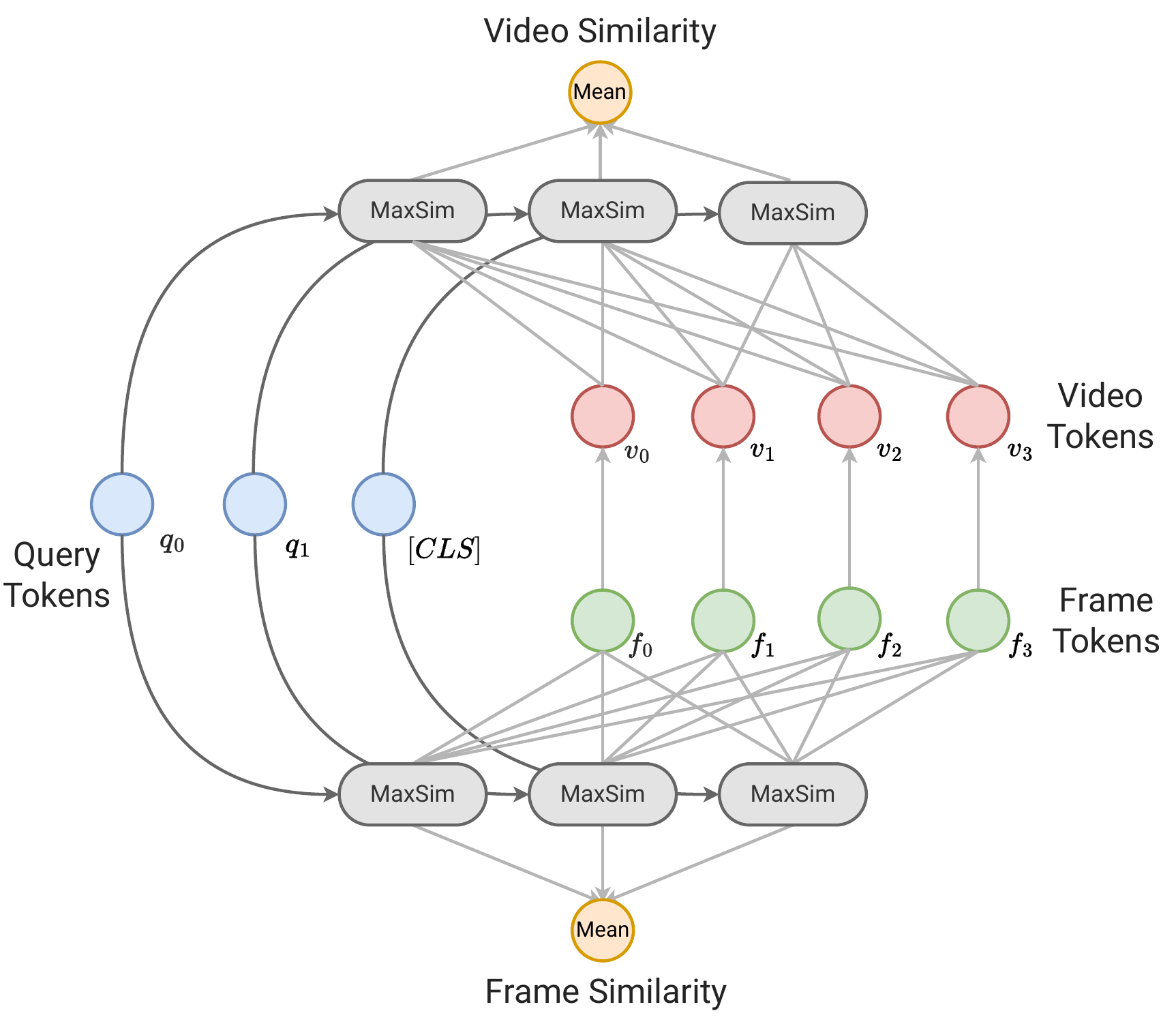}

    \caption{\methodname architecture, combining 
    token-wise interaction on both static frame features (green) and temporally contextualized video features (red).}
   
   \vspace{-1em}
   \label{fig:teaser}
\end{figure}

ColBERT \cite{khattab2020ColBERT} is a text-to-text retrieval model that uses a multi-vector bi-encoder retrieval method. This approach achieves an effective middle ground between expensive cross-encoders and single-vector bi-encoders by enabling late interaction between individual query and document tokens, which maintains the efficiency of simple dot product-based interactions but captures more of the relevant context. Specifically, ColBERT's \textit{MaxSim} operator employs a summation over maximum similarity operations, which allows different aspects of the text query to individually detect relevant content in documents. Additionally, ColBERT introduced unique interactions between query padding tokens and document tokens for \textit{soft query augmentation.}

In this work, we introduce \methodname, a bi-encoder approach to T2VR that leverages tokenwise interaction at both the spatial and spatio-temporal levels. \methodname incorporates a modification to the \textit{MaxSim} operation, \textit{MeanMaxSim} (MMS), which replaces the summation with a mean to better accommodate variable length queries and to control the magnitude of the overall similarity. On top of this modified interaction, \methodname uses two MMS operations over both independent visual frame features and contextualized frame features to strengthen the fine-grained spatial and temporal interaction (\cref{fig:teaser}). This dual MMS interaction is trained with a specialized sigmoid-based loss objective to strengthen the independence and compatibility of the spatial and spatio-temporal representations. We find that the combination of the interactions and the loss function enhances the robustness of each representation resulting in a more resilient fusion during evaluation. 

Our contributions can be summarized as follows:
\begin{enumerate}
    \item We introduce a novel T2VR method, \methodname, a multi-level late-interaction retrieval model that provides better retrieval effectiveness with comparable model size. 
    \item We introduce a richer form of tokenwise interaction by performing MMS at both the spatial and spatio-temporal levels. 
    \item We introduce a sigmoid-based loss for effectively training our version of tokenwise interaction with both spatial and spatio-temporal visual features.
    \item We present a series of experiments that analyze key aspects of our method, including interactions, training objectives and query augmentations.
\end{enumerate}

\section{Related Works}
\label{sec:related}

\paragraph{Text-to-Video Retrieval.}

Early works in text-to-video retrieval made use of pre-trained experts \cite{gabeur2020MMT, liu2020CollaborativeExperts, wang2021T2VLAD} to represent videos. Later attempts to perform end-to-end video-language pre-training \cite{zhu2020ActBERT, xu2021VideoCLIP} (on datasets like HowTo100M \cite{miech2019HowTo100M}) saw limited success due to lack of scale and the poor quality of paired text. Frozen in Time \cite{bain2021FrozenTime} showed that image-text and video-text pairs could be used to train an enhanced bi-encoder model for video retrieval.

Since the advent of the groundbreaking image-text contrastive model CLIP \cite{radford2021CLIP}, several works have sought to adapt it to the video retrieval problem \cite{cheng2021CAMoEDSL, liu2022TS2-Net, wang2022DRL, bain2022CLIPHitchhikersGuideLong, gorti2022X-Pool, zhao2022CenterCLIP, gao2022CLIP2TV, ni2022X-CLIP, luo2022CLIP4Clip, ma2022X-CLIP, fang2023CLIP2Video, xue2023CLIP-ViP}. Many of these methods use CLIP to extract frame- or even patch-level representations, with some employing additional transformer layers for temporal modeling.
Other work has looked to create stronger temporal representations of the video. \citet{liu2022TS2-Net} adopt token shift operations and selection modules to further improve video encoding and increase interaction amongst meaningful frames. \citet{deng2023PromptSwitch} continue to build upon this approach using a ``prompt cube" to force interaction between all pairs of frames in the video which are then aggregated into a final representation.

\vspace{-1.5em}
\paragraph{Tokenwise Late Interaction.}
In the domain of text document retrieval, ColBERT \cite{khattab2020ColBERT} showed the promise of contextualized late interaction techniques, which enable fine-grained interaction between queries and documents while maintaining efficiency at retrieval time. 
FILIP \cite{yao2022FILIP} applies a similar idea to image-text matching, where individual query token features interact with image patch features. 
Recently, ColPali \cite{faysse2024ColPali} applied this to visual document retrieval with vision-language models for retrieving PDF files with textual queries. Likewise, several works also explored tokenwise interaction for video retrieval \cite{wang2022DRL, ma2022X-CLIP, jiang2022TencentText-VideoRetrieval, wu2023Cap4Video, fang2023UATVR, guan2023PIDRo, wang2023UCoFiA}. Among them, X-CLIP \cite{ma2022X-CLIP} uses multi-grained interactions on both the query (word and sentence) and document (frame and video) sides. Another variant, DRL \cite{wang2022DRL}, uses weighted tokenwise interaction to take into account the importance of query words and video frames. UATVR \cite{fang2023UATVR} adds extra learnable tokens as input to the query and video encoders to enrich the tokenwise interaction. Unlike \methodname, none of the aforementioned techniques perform interaction on both spatial and spatio-temporal visual features.
\section{Preliminaries}
\label{sec:preliminaries}

\paragraph{Problem Formulation \& Notation.}
Text-to-video retrieval is the task of ranking a collection of videos based on relevance to a given natural language text query. Formally, given a sequence of text query tokens $Q=\{q_i, \ldots, q_M\}$ and a set of videos $\{V_1,\ldots, V_C\}$, the objective of a T2VR method is to produce a video ranking that aligns with true relevance judgments. In this work, we focus on the T2VR setting where only visual information (\ie video frames) can be used to assess query-video relevance. 

We denote each video as a temporally ordered sequence of sampled frames: $V=\{f_1,\ldots\,f_N\}$, where each $f_i$ has an independent spatial representation $\bfr_i$ (extracted by an image encoder) and temporally contextualized representations $\bv_i$ (produced by a temporal transformer operating on $\{\bfr_1,\ldots,\bfr_N\}$). 

\vspace{-1.5em}
\paragraph{Interaction Mechanisms.}
When considering how query tokens should interact with video frames to compute a query-video relevance score, several options exist. Among the simplest of these are single-vector techniques, like CLIP4Clip \cite{luo2022CLIP4Clip}, which perform a pooling operation over frame features to arrive at a single unified video representation. Similarly, the query can also be represented using a single vector $\bq$, typically using a special aggregation token in a text transformer model. Then, the similarity score can be computed via a dot product (equivalent to cosine similarity, assuming L2 normalization) of the query and video vectors. Formally, the similarity computation using mean pooling (MP) is defined as:

\begin{equation}
    MP(Q, V) = \bq \cdot \frac{1}{N} \sum_{i=1}^N \bfr_i
\end{equation}

While simple and efficient, the MP approach assumes that single vectors are sufficient to adequately represent both the query and video content. Such an assumption may not hold as the complexity of text queries and videos increases. An alternative approach is to employ a more fine-grained interaction between query and video by computing cosine similarity at the individual token level. For example, a ColBERT-style \cite{khattab2020ColBERT} summation over \textit{MaxSim} operations (SMS) could be applied to video retrieval as:

\begin{equation}
    SMS(Q, V) =  \sum_{j=1}^M \max_{i} (\bq_j \cdot \bfr_i)
    \label{eqn:summaxsim}
\end{equation}

The SMS formulation for similarity calculation allows each query token representation to effectively ``scan" the video frames for relevant content, then contribute to the summation the maximum similarity found among all frames. We employ a similar fine-grained interaction approach in \methodname.

\section{Method}
\label{sec:method}
\begin{figure*}[t]
  \centering
    \includegraphics[width=\linewidth]{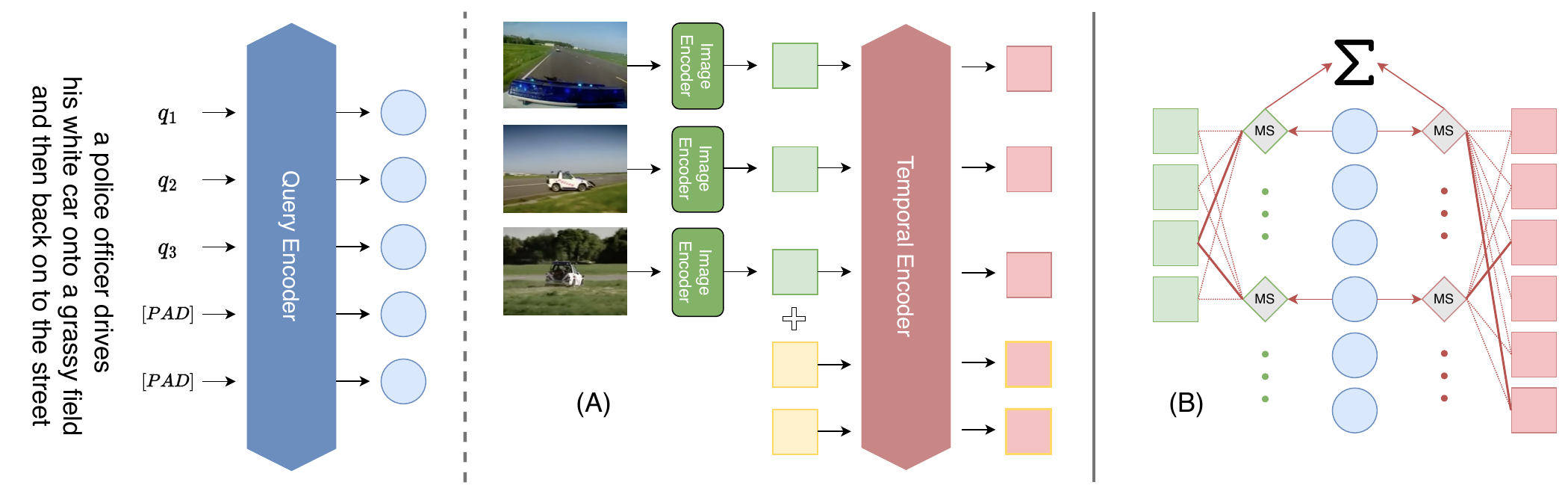}
\vspace{-6mm}
   \caption{Overview of \methodname. (A) Shows the \methodname bi-encoder process. The query is encoded by a text encoder (\lineimg{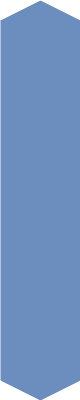}). The images are encoded independently with an image-encoder (\lineimg{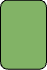}) and produce [CLS] tokens for frames (\lineimg{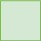}). The frames and additional visual expansion tokens (\lineimg{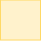}) are passed through the temporal transformer(\lineimg{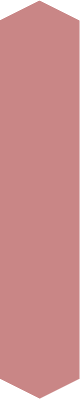}). (B) Shows the dual $\text{MMS}_{FV}$ interaction, where encoded query tokens (\lineimg{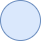}) perform \textit{MaxSim} operations (\lineimg{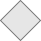}) with the frames (\lineimg{x_fig/inline_graphics/image_token.png}) and video features (\lineimg{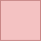}). A mean is then performed over each \textit{MaxSim} for the frames and video features, then the two means are summed to compute the final relevance score.}
   \label{fig:method}
    \vspace{-1.5em}
\end{figure*}

We now describe \methodname, a fine-grained approach for adapting image-text dual encoder models (like CLIP \cite{radford2021CLIP} and SigLIP \cite{zhai2023SigLIP}) for T2VR.
\methodname (depicted in \cref{fig:method}) has 3 main aspects: (i) fine-grained spatial and temporal interaction, performing MMS on both independent frames and their contextualized representations, (ii) query and visual expansion tokens which allow for additional information to be encoded for abstract queries and for additional high-level temporal information from the video, and (iii) a dual sigmoid loss for training strong independent, yet compatible, spatial and spatio-temporal representations.

\subsection{Fine-Grained Spatial \& Temporal Interaction}

The first component of our query-video interaction mechanism is a modified form of SMS (\cref{eqn:summaxsim}), which replaces the summation with a mean to better accommodate variable length queries. This interaction, which we denote as $\text{MMS}_{F}$, operates on static frame features extracted by an image encoder (specifically, using the [CLS] token of a vision transformer \cite{dosovitskiy2021ImageWorth16x16Words}):

\begin{equation}
    MMS_F(Q, V) =  \frac{1}{M} \sum_{j=1}^M \max_{i} (\bq_j \cdot \bfr_i)
\end{equation}
where $\bq_j$ denotes the output of the $j$-th query token from the query encoder (\ie a contextualized query token feature). Because a set of individual image features is unable to capture relationships across time, we also perform temporal modeling by processing the frame [CLS] tokens with additional transformer layers. The output of these temporal transformer layers is a sequence of temporally contextualized ``video" features $\{\bv_1,\ldots,\bv_N\}$. Our method performs tokenwise query interaction with these features as well:

\begin{equation}
    MMS_V(Q, V) =  \frac{1}{M} \sum_{j=1}^M \max_{i} (\bq_j \cdot \bv_i)
\end{equation}

In contrast to \cite{yao2022FILIP, wang2022DRL}, we intentionally calculate both $\text{MMS}_{F}$  and $\text{MMS}_{V}$ in only one direction. In other words, only query token features are used to select relevant visual features and not the other way around.  We argue that there exists an inherent asymmetry between queries and videos, and that the overall similarity score should not be diminished by the presence of video frames that do not correspond with any query tokens. The final query-video similarity score in \methodname is a sum of the frame-level and video-level MMS scores:
\begin{equation}
    MMS_{FV}(Q, {V}) =  MMS_F(Q, V) + MMS_V(Q, V)
\end{equation}
Such summation can also be thought of as the Borda score of the set of frame scores~\cite{darmann2019using}.

The aim of $\text{MMS}_{FV}$ is to better capture the interaction between purely spatial information and spatio-temporal video features from the query by incorporating two levels of interaction. The $\text{MMS}_{F}$ operation locates relevant static information, while the $\text{MMS}_{V}$ operation matches dynamic concepts. 
Unlike previous works that only use features after temporal modeling, the contextualized video representations in \methodname can encode more temporal information because the temporal layers have less need to preserve purely spatial concepts and can instead focus on capturing higher-level cross-frame and global interactions.

\subsection{Query \& Visual Expansion}

In addition to our interaction mechanism, we again take inspiration from ColBERT \cite{khattab2020ColBERT} in our use of \textit{soft query augmentation}. ColBERT finds that including extra padding token features in the tokenwise interaction enhances retrieval performance. The authors hypothesize that augmenting the query with these additional tokens enables a learnable form of query expansion \cite{khattab2020ColBERT, formal2021white}, whereby additional ``search terms" implied by the base query can be computed to enhance retrieval. Following this intuition, we incorporate pad tokens into both the $\text{MMS}_F$ and $\text{MMS}_V$ interactions.

The intuition behind query expansion tokens also extends to the video side. Following \citet{fang2023UATVR}, we incorporate visual expansion tokens, alongside the frame [CLS] tokens, as input to the temporal transformer. 

\subsection{Dual Sigmoid Loss}

Prior methods that train bi-encoder models for T2VR primarily utilize a bi-directional, softmax-based InfoNCE loss \cite{oord2019InfoNCE}, defined as follows:
\[
- \frac{1}{2|B|} \sum_{i=1}^{|B|} \left( \log \frac{e^{t \mathbf{x}_i \cdot \mathbf{y}_i}}{\sum_{j=1}^{|B|} e^{t \mathbf{x}_i \cdot \mathbf{y}_j}} + \log \frac{e^{t \mathbf{x}_i \cdot \mathbf{y}_i}}{\sum_{j=1}^{|B|} e^{t \mathbf{x}_j \cdot \mathbf{y}_i}} \right)
\]

In the context of text-video retrieval, $\mathbf{x}$ and $\mathbf{y}$ would be text and video representations (with $B$ denoting batch size and $t$ a learnable logit scaling factor). The overall loss incorporates both text-to-video and video-to-text InfoNCE losses by performing softmax normalization across both dimensions of the batch similarity matrix. However, InfoNCE is sensitive to the negative example selection~
\cite{chuang_debiased_2020, shen_simple_2020, karpukhin2020dense}.
Such examples may be hard to obtain without the help of an effective matching or selection model to start with, which is the kind of model that we aim to train in the first place. 
Furthermore, recent advancements in image-text contrastive learning have demonstrated the advantages of sigmoid-based losses over their softmax counterparts \cite{zhai2023SigLIP}. 
The sigmoid loss turns the original contrastive objective into a series of independent binary classification tasks, thereby eliminating the need for computing global normalization factors. 
 
The sigmoid loss has also been shown to be more robust to noisy data~\cite{zhai2023SigLIP}, which is prevalent in T2VR datasets in both the quality of annotations \cite{chen2024MSRVideoTextDataset} and the ambiguity of abstract text queries and descriptions \cite{zhang_video-aided_2021}.

For these reasons, we adopt the sigmoid loss when training \methodname. Specifically, the loss is defined as: 
%
\begin{equation}
    - \frac{1}{|B|}\sum_{i=1}^{|B|}\sum_{j=1}^{|B|} \log\frac{1}{1+e^{z_{ij}(-t\cdot MMS(Q_i,V_j) + b)}}
\end{equation}
%
%
where $z_{ij}$ is a label indicating positive ($+1$) and negative ($-1$) pairings, $t$ denotes a learnable logit scaling factor and $b$ is a learnable logit bias.

Building upon the sigmoid loss and the fine-grained spatial and spatio-temporal interactions in $\text{MMS}_{FV}$, 
we propose a dual loss function that fuses $\text{MMS}_F$ and $\text{MMS}_V$ at the ranking level. 
Since the information in $\text{MMS}_F$ and $\text{MMS}_V$ propagates from different levels, their magnitudes will naturally differ. 
Given this, it is preferable to compute separate sigmoid losses on the $\text{MMS}_F$ and $\text{MMS}_V$ similarity matrices and avoid the multi-loss scaling issue~\cite{li2023MultipleTasksMultipleObjectives, cipolla2018MultitaskLearningUsing}.
Furthermore, separate losses encourage stronger independent representations from each level (frame and video). Thus, we employ a dual loss formulation (\cref{eqn:dual_sigmoid_loss}) that computes the global loss as the linear combination of the losses of the $\text{MMS}_F$ and $\text{MMS}_V$ interactions.

\begin{align*}
    \loss_F =  - \frac{1}{|B|}\sum_{i=1}^{|B|}\sum_{j=1}^{|B|} \log\frac{1}{1+e^{z_{ij}(-t\cdot MMS_F(Q_i,V_j) + b)}} \\
    \loss_F =  - \frac{1}{|B|}\sum_{i=1}^{|B|}\sum_{j=1}^{|B|} \log\frac{1}{1+e^{z_{ij}(-t\cdot MMS_V(Q_i,V_j) + b)}}
\end{align*}

\begin{equation}
    \loss_D = \lambda_F \loss_F + \lambda_V \loss_V
\label{eqn:dual_sigmoid_loss}
\end{equation}

\noindent $\lambda_F$ and $\lambda_V$ are additional hyperparameters, allowing more importance to be placed on spatial or temporal features. 

\begin{table*}[ht!]
\centering
\resizebox{0.93\textwidth}{!}{
\begin{tabular}{lcccc|cccc|cccc}
\hline

\hline

\hline\\[-3mm]
 \multicolumn{1}{l}{\multirow{2}{*}{\textbf{Method}}} & \multicolumn{4}{c|}{\textbf{MSR-VTT}} & \multicolumn{4}{c|}{\textbf{MSVD}} & \multicolumn{4}{c}{\textbf{VATEX}}\\  
\multicolumn{1}{c}{} & {R@1} & {R@5} & {R@10} & {nDCG} & {R@1} & {R@5} & {R@10} & {nDCG} & {R@1} & {R@5} & {R@10} & {nDCG} \\


\hline
ClipBERT~\cite{lei2021ClipBERT}                    
    & 22.0 & 46.8 & 59.9 & $-$    
    & $-$ & $-$ & $-$ & $-$       
    & $-$ & $-$ & $-$ & $-$ \\
Support Set~\cite{patrick2020support}                 
    & 30.1 & 58.5 & 69.3 & $-$     
    & 28.4 & 60.0 & 72.9 & $-$    
    & 45.9 & 82.4 & 90.4 & $-$\\
Frozen~\cite{bain2021FrozenTime}                      
    & 32.5 & 61.5 & 71.2 & $-$     
    & 33.7 & 64.7 & 76.3 & $-$    
    & $-$ & $-$ & $-$ & $-$\\
\hline
\rowcolor{gray!9} \multicolumn{13}{c}{\textit{ViT-B/32}} \\
CLIP4Clip-meanP~\cite{luo2022CLIP4Clip}             
    & 43.1 & 70.4 & 80.8 & $-$          
    & 46.2 & 76.1 & 84.6 & $-$     
    & $-$ & $-$ & $-$ & $-$\\
CLIP4Clip-seqTransf~\cite{luo2022CLIP4Clip}         
    & 44.5 & 71.4 & 81.6 & $-$   
    & 45.2 & 75.5 & 84.3 & $-$    
    & $-$ & $-$ & $-$ & $-$\\
CenterCLIP~\cite{zhao2022CenterCLIP}                  
    & 44.2 & 71.6 & 82.1 & $-$   
    & \underline{47.6} & 77.5 & \underline{86.0} & $-$  
    & $-$ & $-$ & $-$ & $-$\\
CLIP2TV~\cite{gao2022CLIP2TV}                     
    & 46.1 & 72.5 & 82.9 & $-$    
    & 47.0 & \underline{76.5} & 85.1 & $-$    
    & $-$ & $-$ & $-$ & $-$\\
TS2-Net~{\cite{liu2022TS2-Net}}     
    & 47.0 & 74.5 & \underline{83.8} & $-$                  
    & 44.6 & 75.8 & $-$ & $-$        
    & 59.1 & 90.0 & 95.2  & $-$\\
X-CLIP~{\cite{ma2022X-CLIP}}     
    & 46.1 & 73.0 & $-$ & $-$                  
    & 47.1 & 77.8 & $-$ & $-$        
    & $-$ & $-$ & $-$  & $-$\\
DRL~\cite{wang2022DRL}                         
    & \underline{47.4} & \underline{74.6} & \underline{83.8} & $-$    
    & \textbf{48.3} & \textbf{79.1} & \textbf{87.3} & $-$  
    &\textbf{63.5} & \textbf{91.7} & \textbf{96.5} & $-$\\

\rowcolor{green!7} 
\methodname (CLIP-B/32)             
    & \textbf{48.1} & \textbf{74.9}  & \textbf{83.9} & \textbf{0.652} 
    & 46.0 & 75.0  & 84.0 & \textbf{0.645}    
    &  \underline{61.8} & \underline{90.8}  & \underline{95.7} & \textbf{0.794}  \\
\hline

\rowcolor{gray!9} \multicolumn{13}{c}{\textit{ViT-B/16}} \\
CLIP2TV~\cite{gao2022CLIP2TV}                     
    & 49.3 & 74.7 & 83.6 & $-$       
    & $-$ & $-$ & $-$ & $-$                        
    & $-$ & $-$ & $-$ & $-$\\
TS2-Net~\cite{liu2022TS2-Net} 
    & 49.4 & 75.6 & \underline{85.3} & $-$
    & $-$ & $-$ & $-$ & $-$ 
    & $-$ & $-$ & $-$ & $-$\\

X-CLIP~\cite{ma2022X-CLIP}          
    & 49.3 & 75.8 & 84.8 & $-$    
    & \underline{50.4} & 80.6 & $-$ & $-$    
    & $-$ & $-$ & $-$ & $-$\\
DRL~\cite{wang2022DRL}                         
    & 50.2& \underline{76.5} & 84.7 & $-$    
    & 50.0 & \underline{81.5} & \textbf{89.5} & $-$    
    & 65.7 & 92.6 & 96.7 & $-$\\
SigLIP4CLIP-meanP \cite{luo2022CLIP4Clip}          
    & 46.2 & 71.9 & 81.6 & \underline{0.633}   
    & 50.1 & 78.3 & 86.7 & 0.680 
    & 64.2 & 92.2 & 96.4 & 0.812 \\
SigLIP4CLIP-seqTransf \cite{luo2022CLIP4Clip}          
    & 45.7 & 71.0 & 80.0 & 0.623 
    & 47.4 & 76.6 & 85.0 & 0.659 
    & 66.1 & \underline{92.9} & \underline{96.8} & 0.824 \\
\rowcolor{green!7} 
\methodname(CLIP-B/16)              
    & \underline{51.0} & \textbf{77.1} & \textbf{85.5} & \textbf{0.677}    
    & 50.2 & 79.6 & 87.8 & \underline{0.683}
    & \underline{66.8} & \underline{92.9} & \underline{96.8} & \underline{0.826} \\
\rowcolor{green!7} 
\methodname(SigLIP-B/16) 
    & \textbf{51.5} & 76.3          & \textbf{85.5} & \textbf{0.677}  
    & \textbf{55.2} & \textbf{82.9} & \underline{89.4} & \textbf{0.724}    
    & \textbf{68.0} & \textbf{93.4} & \textbf{96.9} & \textbf{0.833} \\
\hline

\textcolor{lightgray}{Cap4Video \cite{wu2023Cap4Video}}
    & \textcolor{lightgray}{51.4} & \textcolor{lightgray}{75.7} & \textcolor{lightgray}{83.9} & \textcolor{lightgray}{$-$} 
    & \textcolor{lightgray}{51.8} & \textcolor{lightgray}{80.8} & \textcolor{lightgray}{88.3} & \textcolor{lightgray}{$-$} 
    & \textcolor{lightgray}{66.6} & \textcolor{lightgray}{93.1} & \textcolor{lightgray}{97.0} & \textcolor{lightgray}{$-$} \\
\textcolor{lightgray}{X-Pool \cite{gorti2022X-Pool}}
    & \textcolor{lightgray}{46.9} & \textcolor{lightgray}{72.8} & \textcolor{lightgray}{82.2} & \textcolor{lightgray}{$-$} 
    & \textcolor{lightgray}{47.2} & \textcolor{lightgray}{77.4} & \textcolor{lightgray}{86.0} & \textcolor{lightgray}{$-$}
    & \textcolor{lightgray}{$-$} & \textcolor{lightgray}{$-$} & \textcolor{lightgray}{$-$} & \textcolor{lightgray}{$-$} \\
\textcolor{lightgray}{InternVideo2-6B \cite{wang2024InternVideo2}}
    & \textcolor{lightgray}{55.9} & \textcolor{lightgray}{78.3} & \textcolor{lightgray}{85.1} & \textcolor{lightgray}{$-$} 
    & \textcolor{lightgray}{59.3} & \textcolor{lightgray}{84.4} & \textcolor{lightgray}{89.6} & \textcolor{lightgray}{$-$} 
    & \textcolor{lightgray}{71.5} & \textcolor{lightgray}{94.0} & \textcolor{lightgray}{97.1} & \textcolor{lightgray}{$-$} \\
\hline

\hline

\hline
\end{tabular}
}

\vspace{-0.5em}
\caption{Results on sentence-to-video retrieval tasks using MSR-VTT, MSVD and VATEX datasets. \textbf{Bold} indicates best performance for a particular model size, and \underline{underline} indicates second best.}
\label{tab:msrvtt_msvd_vatex_results}
\end{table*}

\begin{table*}[ht!]
\centering
\resizebox{0.7\textwidth}{!}{
\begin{tabular}{lcccc|cccc}
\hline

\hline

\hline\\[-3mm]
 \multicolumn{1}{l}{\multirow{2}{*}{\textbf{Method}}} & \multicolumn{4}{c|}{\textbf{DiDeMo}} & \multicolumn{4}{c}{\textbf{ActivityNet}}\\  
\multicolumn{1}{c}{} & {R@1} & {R@5} & {R@10} & {nDCG} & {R@1} & {R@5} & {R@10} & {nDCG} \\


\hline
ClipBERT~\cite{lei2021ClipBERT}         
    & 20.4 & 48.0 & 60.8 & $-$    
    & 21.3 & 49.0 & 63.5 & $-$ \\
All-in-One~\cite{wang2023all}       
    & 32.7 & 61.4 & 73.5 & $-$    
    & 22.4 & 53.7 & 67.7 & $-$ \\
Frozen~\cite{bain2021FrozenTime}        
    & 34.6 & 65.0 & 74.7 & $-$    
    & $-$ & $-$ & $-$ & $-$  \\
\hline
\rowcolor{gray!9} \multicolumn{9}{c}{\textit{ViT-B/32}} \\
CLIP4Clip-meanP~\cite{luo2022CLIP4Clip} 
    & 43.4 & 70.2 & 80.6 & $-$    
    & 40.5 & 72.4 & $-$ & $-$  \\
CLIP4Clip-seqTransf~\cite{luo2022CLIP4Clip} 
    & 43.4 & 70.2 & 80.6 & $-$    
    & 40.5 & 72.4 & $-$  & $-$  \\
CenterCLIP~\cite{zhao2022CenterCLIP}  
    & $-$ & $-$ & $-$ & $-$      
    & 43.9 & 74.6 & \underline{85.8} & $-$    \\
CLIP2TV~\cite{gao2022CLIP2TV}       
    & 45.5 & 69.7 & 80.6 & $-$    
    & 44.1 & \textbf{75.2} & $-$ & $-$    \\
X-CLIP~\cite{ma2022X-CLIP}  
    & 45.2 & \underline{74.0} & $-$ & $-$    
    & \underline{44.3} & 74.1 & $-$ & $-$ \\
DRL~\cite{wang2022DRL}     
    & \underline{47.9} & 73.8 & \underline{82.7} & $-$    
    & 44.2 & 74.5 & \textbf{86.1} & $-$  \\

\rowcolor{green!7} 
\methodname (CLIP-B/32)  
    & \textbf{48.2} & \textbf{75.1} & \textbf{83.7} & \textbf{0.654}                 
    & \textbf{45.5} & \underline{74.6} & 85.5 & \textbf{0.645}\\
\hline

\rowcolor{gray!9} \multicolumn{9}{c}{\textit{ViT-B/16}} \\
CLIP4Clip-meanP~\cite{luo2022CLIP4Clip} 
    & 44.8 & 75.1 & $-$ & $-$    
    & 44.0 & 73.9 & $-$ & $-$  \\
CLIP4Clip-seqTransf~\cite{luo2022CLIP4Clip} 
    & 44.8 & 73.4 & $-$ & $-$    
    & 44.5 & 75.2 & $-$ & $-$  \\
X-CLIP~\cite{ma2022X-CLIP}  
    & 47.8 & \textbf{79.3} & $-$ & $-$    
    & \underline{46.2} & 75.5 & $-$ & $-$ \\
DRL~\cite{wang2022DRL} 
    & 49.0 & 76.5 & 84.5 & $-$    
    & \underline{46.2} & \underline{77.3} & \textbf{88.2} & $-$  \\

\rowcolor{green!7} 
\methodname(CLIP-B/16)      
    & \textbf{51.9} & \underline{78.3} & \textbf{85.6} & \textbf{0.682}
    & \textbf{50.6} & \textbf{78.0} & \underline{87.9} & \textbf{0.685}\\
\rowcolor{green!7} 
\methodname(SigLIP-B/16)    
    & \underline{51.7} & 76.1 & \underline{84.8} & \underline{0.675}    
    & 45.8 & 76.3 & 86.7 & \underline{0.656}\\
\hline

\textcolor{lightgray}{InternVideo2-6B \cite{wang2024InternVideo2}}
    & \textcolor{lightgray}{57.9} & \textcolor{lightgray}{80.0} & \textcolor{lightgray}{84.6} & \textcolor{lightgray}{$-$} 
    & \textcolor{lightgray}{63.2} & \textcolor{lightgray}{85.6} & \textcolor{lightgray}{92.5} & \textcolor{lightgray}{$-$} \\
\hline

\hline

\hline
\end{tabular}
}

\vspace{-0.5em}
\caption{Results on paragraph-to-video retrieval tasks using DiDeMo and ActivityNet datasets. \textbf{Bold} indicates best performance for a particular model size, and \underline{underline} indicates second best.}
\label{tab:didemo_activitynet_results}
\end{table*}

\section{Experiments}
\label{sec:experiments}
\subsection{Datasets}
We evaluate \methodname on several T2VR datasets, in which video captions serve as proxies for user queries. We use only English captions for all datasets.
\begin{itemize}
    \item \textit{MSR-VTT} \cite{xu2016MSR-VTT} contains 10,000 total videos, each paired with 20 captions. We use the \texttt{Training-9K} and \texttt{1K-A} splits for training and testing respectively.
    \item \textit{MSVD} \cite{chen2011MSVD} contains 1,200 training videos and 670 test videos, each paired with roughly 40 captions.
    \item \textit{VATEX} \cite{wang2019VaTeX} contains 25,991 training videos and 1,500 test videos, each with 10 captions.
    \item \textit{DiDeMo} \cite{hendricks2017DiDeMo} contains 8,391 training videos and 1,004 test videos. Each video is paired with approximately four temporally localized captions, which we concatenate to create a paragraph-video retrieval task.
    \item \textit{ActivityNet} \cite{heilbron2015ActivityNet} contains 10,009 training videos and 4,917 test videos. We again concatenate the temporally localized captions to create a paragraph-video retrieval task for this dataset.
\end{itemize}

\subsection{Implementation Details}

\paragraph{Network Architecture \& Training.}
Both the query and video encoders in \methodname are initialized from CLIP ViT-B/16 or CLIP ViT-B/32. We additionally introduce a variant of the ViT-B/16 initialized from SigLIP \cite{zhai2023SigLIP} which was trained on the WebLI dataset using a sigmoid loss. For temporal modeling, we use 4 transformer layers based on the text encoder of the underlying dual encoder model. We set the number of visual expansion tokens to 2, and set both $\lambda_F$ and $\lambda_V$ to 1 during training. Our models are fine-tuned using the Adam \cite{kingma2015Adam} optimizer, with learning rate of $1\times10^{-7}$ for pre-trained image and text encoder parameters and $1\times10^{-4}$ for temporal transformer parameters. We freeze the positional encodings, patch embeddings and token embeddings during fine-tuning. All other transformer parameters are trained. More details about training settings and computational burden can be found in the Appendix.

\vspace{-1.5em}
\paragraph{Text Pre-Processing.}
We adopt the same tokenizer and special token mappings used in the original CLIP and SigLIP models. For CLIP \cite{radford2021CLIP}, we prepend the token sequence with a \code{<|startoftext|>} token and add an \code{<|endoftext|>} token at the end. For query augmentation using CLIP, we enable self-attention and MMS interaction with additional pad tokens (ID \#0, which corresponds to an exclamation point \code{!}) used to fix the token sequence length to 32 for MSR-VTT, MSVD and VATEX, and 64 for DiDeMo and ActivityNet. Because SigLIP uses a ``last token" aggregation strategy and always performs self-attention across a length-64 padded token sequence, we use 64 text tokens when using SigLIP as a backbone.

\vspace{-1.5em}
\paragraph{Video Pre-Processing.}
We perform TSN-style \cite{wang2016TemporalSegmentNetworks} uniform sampling to select frames from videos. 
Each frame is resized to $224 \times 224$ without maintaining aspect ratio in order to avoid information loss resulting from center cropping.
In line with previous work, we sample 12 frames for MSR-VTT, MSVD and VATEX, while using 64 frames for DiDeMo and ActivityNet.

\subsection{Baselines \& Evaluation Metrics}
Our comparisons to other work focus on alternative bi-encoder approaches for text-video retrieval. We compare against alternative interaction mechanisms and context aggregation strategies rather than orthogonal approaches like captioning \cite{wu2023Cap4Video}, large-scale video-text pre-training \cite{rizve2024VidLA}, and expensive methods that use cross-modal transformers or multimodal large language models \cite{cao2024RAP, chen2023VAST, gorti2022X-Pool, wang2024InternVideo2}. In our comparisons, we include the best bi-encoder approaches for text-video retrieval \cite{gao2022CLIP2TV, liu2022TS2-Net, wang2022DRL, zhao2022CenterCLIP, luo2022CLIP4Clip, ma2022X-CLIP} built upon CLIP-B/32, CLIP-B/16 and SigLIP-B/16.

For a fairer comparison with our SigLIP encoder variant, we also upgrade the backbone in CLIP4CLIP \cite{luo2022CLIP4Clip} to create SigLIP4Clip. We implement the mean pooling (meanP) and sequence transformer (seqTransf) variants of the CLIP4Clip method and perform fine-tuning using an InfoNCE loss. 

For evaluation, we report recall at 1 (R@1), 5 (R@5) and 10 (R@10). As advocated for by \cite{wray2021SemanticSimilarityVideo}, we also include normalized discounted cumulative gain (nDCG), a commonly used metric in text-based information retrieval. We use nDCG@10 and abbreviate it as nDCG in our tables. For all metrics, higher number indicate better performance. 

\subsection{Benchmark Results}

In \cref{tab:msrvtt_msvd_vatex_results}, we show results on three sentence-to-video retrieval datasets. We see that \methodname achieves competitive or state-of-the-art results in several settings. When using CLIP-B/32 as a backbone, we find that \methodname outperforms other approaches that utilize tokenwise interaction (even those that use more granular patch-level information \cite{liu2022TS2-Net}). We find that DRL \cite{wang2022DRL}, when using the cheaper CLIP-B/32 model, outperforms \methodname on MSVD and VATEX, likely due to its use of channel decorrelation regularization and extra learnable token weightings. When using ViT-B/16 backbones, \methodname exhibits even more impressive retrieval performance compared to alternative methods. For example, with SigLIP-B/16, \methodname sets a new state-of-the-art on MSRVTT, MSVD and VATEX. The results using CLIP4Clip with an upgraded SigLIP model indicate that our performance gains are not solely attributed to the improved backbone, but rather that our two-level tokenwise interaction strategy provides an effective way to match fine-grained text and video features.

In \cref{tab:didemo_activitynet_results} we report results on two paragraph-to-video retrieval benchmarks. We find again that \methodname achieves impressive results, with particularly strong performance on DiDeMo using all three backbones. While our CLIP-B/16-based model outperforms other methods on ActivityNet by a large margin, we find that the SigLIP-based variant of \methodname performs relatively poorly on ActivityNet. This is likely because SigLIP was pre-trained on text with a maximum length of 16 tokens, while the captions in ActivityNet are of much longer length. 

Unless otherwise indicated, all results are reported without use of DSL \cite{cheng2021CAMoEDSL} or QB-Norm \cite{bogolin2022QB-Norm}.

\section{Additional Analysis \& Discussion}
\label{sec:analysis}
We perform additional analysis on each modeling choice made in \methodname using MSR-VTT~\cite{xu2016MSR-VTT}. 

\vspace{-1.5em}
\paragraph{Interaction Mechanisms.}
\begin{table}[t]
\begin{center}
\resizebox{0.999\linewidth}{!}{
\begin{tabular}{c|c|cc|cccc} 
\toprule
\textbf{Type} & \textbf{Name} & \textbf{Frame} & \textbf{Video} & {R@1} & {R@5} & {R@10} & {nDCG} \\ 
\midrule
    \multirow{2}{*}{MP}     & $-$ & \cmark & \xmark    & 42.1 & 69.6 & 79.3 & 0.601  \\
                            & $-$ & \xmark & \cmark    & 43.4 & 70.4 & 80.1 & 0.610   \\ 
\midrule
\multirow{4}{*}{MMS}
                    & $\text{MMS}_F$   & \cmark & \xmark   & 44.3     & 71.5    & 82.4 & 0.626 \\ 
                    & $\text{MMS}_V$   & \xmark & \cmark   & 47.0     & 74.1    & 82.4 & 0.643 \\ 
  & \cellcolor{green!7} $\text{MMS}_{FV}$ & \cellcolor{green!7}\cmark & \cellcolor{green!7}\cmark   & \cellcolor{green!7}\textbf{48.1} & \cellcolor{green!7}\textbf{74.9}  & \cellcolor{green!7}\textbf{83.9} & \cellcolor{green!7}\textbf{0.652} \\ 
                    &  RRF        & \cmark & \cmark   & 46.8    & 74.6     & 83.8  & 0.646 \\ 
\bottomrule
\end{tabular}}
\end{center}
\vspace{-6mm}
\caption{Effect of interaction type (coarse- and fine-grained) and interaction involvement (frame and video features) on MSR-VTT retrieval. All results use the CLIP-B/32 backbone.}
\vspace{-0.5em}
\label{tab:interactions}
\end{table}

In \cref{tab:interactions}, we compare strategies for interacting query token features with visual features. We observe that mean pool (MP), a coarse-grained similarity computation between the query [CLS] token and a mean pooling across frame features, results in significantly 
lower retrieval performance 
than fine-grained tokenwise interaction methods. In fact, we find that MMS operating only on static frame features ($\text{MMS}_F$) outperforms MP with additional temporal modeling. Employing MMS on temporally contextualized visual features ($\text{MMS}_V$) leads to large performance gains over the frame-only approach. Finally, we see the best overall performance when combining both frame and video-level tokenwise similarities in $\text{MMS}_{FV}$, suggesting that the two metrics are complementary, enabling specialization on different concepts.

Given that $\text{MMS}_F$ and $\text{MMS}_V$ produce independent, yet complementary, similarity scores for each video, we consider an alternative strategy for combining their rankings rather than adding them together as in $\text{MMS}_{FV}$. Specifically, we explore the use of reciprocal rank fusion (RRF) \cite{cormack2009reciprocal}, a widely used method for combining the outputs of multiple ranking methods in document retrieval. RRF is particularly advantageous when the similarity scores produced by different methods exist on different scales, which is possible when computing similarity scores before and after a temporal transformer. Interestingly, we find that simply summing the frame and video MMS scores ($\text{MMS}_{FV}$) results in superior retrieval performance compared to using RRF in evaluation. 
This observation suggests that the actual score differences in MMS$_F$ and MMS$_V$ are meaningful, and that fusing only with the reciprocal of ranks disposes of useful information, indicating potential opportunities to improve the performance even further through more score alignment between the two scoring functions. 

\vspace{-1.5em}
\paragraph{Choice of Loss Function.}
\begin{table}[t]
\begin{center}
\resizebox{0.95\linewidth}{!}{
\begin{tabular}{c|ccccc} 
\toprule
\textbf{Loss Type} & \textbf{Loss Function} & {R@1} & {R@5} & {R@10} & {nDCG} \\ 
\midrule
\multirow{2}{*}{Combined}   
    & InfoNCE                           & 45.1    & 71.4   & 81.4 & 0.625 \\ 
    & Sigmoid                           & 47.1    & 73.1   & 83.9 & 0.647 \\
\midrule
\multirow{2}{*}{Dual} 
    & InfoNCE                           & 45.3     & 73.0           & 83.8  & 0.640\\ 
    & \cellcolor{green!7} Sigmoid       & \cellcolor{green!7}\textbf{48.1}   & \cellcolor{green!7}\textbf{74.9}  & \cellcolor{green!7}\textbf{83.9} & \cellcolor{green!7}\textbf{0.652} \\
\bottomrule
\end{tabular}}
\end{center}
\vspace{-6mm}
\caption{Effect of loss type and loss function on MSR-VTT retrieval. All results use the a CLIP-B/32 backbone.}
\vspace{-1.2em}
\label{tab:loss_functions}
\end{table}

In \cref{tab:loss_functions}, we analyze the effect of different loss functions on the retrieval performance of \methodname. We find that the sigmoid loss \cite{zhai2023SigLIP} significantly outperforms the standard InfoNCE \cite{oord2019InfoNCE} contrastive loss typically used in T2VR. Despite not being pre-trained with sigmoid loss, CLIP-B/32 still benefits from its use during fine-tuning. We find that the choice of logit scale and bias parameters is critical for proper convergence using a sigmoid loss. We fix these values to those obtained from SigLIP pre-training ($t=4.77$ and $b=-12.93$).

We also compare our dual sigmoid loss formulation, in which two similarity matrices are created using $\text{MMS}_F$ and $\text{MMS}_V$ scores, with a ``combined" alternative where only a single similarity matrix is created during training by summing the frame-level and video-level matrices:
\vspace{-2mm}
\setlength{\abovedisplayskip}{0pt}
\setlength{\abovedisplayshortskip}{0pt}

\begin{equation*}
    \loss_{\text{combined}} = - \frac{1}{|B|}\sum_{i=1}^{|B|}\sum_{j=1}^{|B|} \log\frac{1}{1+e^{z_{ij}(-t\cdot MMS_{FV}+ b)}}
\end{equation*}
\vspace{-4mm}

We observe a small improvement when using the dual loss over the combined method, possibly due to stronger learning objectives on the individual MMS interactions.

\vspace{-1.5em}
\paragraph{Query Augmentation.}
In \cref{tab:query_padding} we analyze the effect of applying ColBERT's \cite{khattab2020ColBERT} soft query augmentation technique in the video retrieval setting. For both CLIP and SigLIP we observe improvements in retrieval performance when query padding tokens participate in the tokenwise MMS interactions. The increase is more substantial when using CLIP, likely due to the attention mechanism used in the CLIP text encoder. CLIP's text encoder employs a causal attention mask, which means that earlier tokens in the sequence are not permitted to attend to future ones, effectively limiting their contextualization and thus their utility in tokenwise interaction. Query augmentation can mitigate this effect by adding additional tokens to the interaction that are able to attend to all of the original query tokens. SigLIP, on the other hand, uses full (\ie bidirectional) self-attention in its text encoder. Since its text tokens are already fully contextualized, the benefit of the additional tokens is more limited. However, the positive improvement gives credence to \citet{khattab2020ColBERT}'s hypothesis that query augmentation can introduce additional soft search terms that enhance retrieval. 

\begin{figure}[t]
  \centering
    \includegraphics[width=0.99\linewidth]{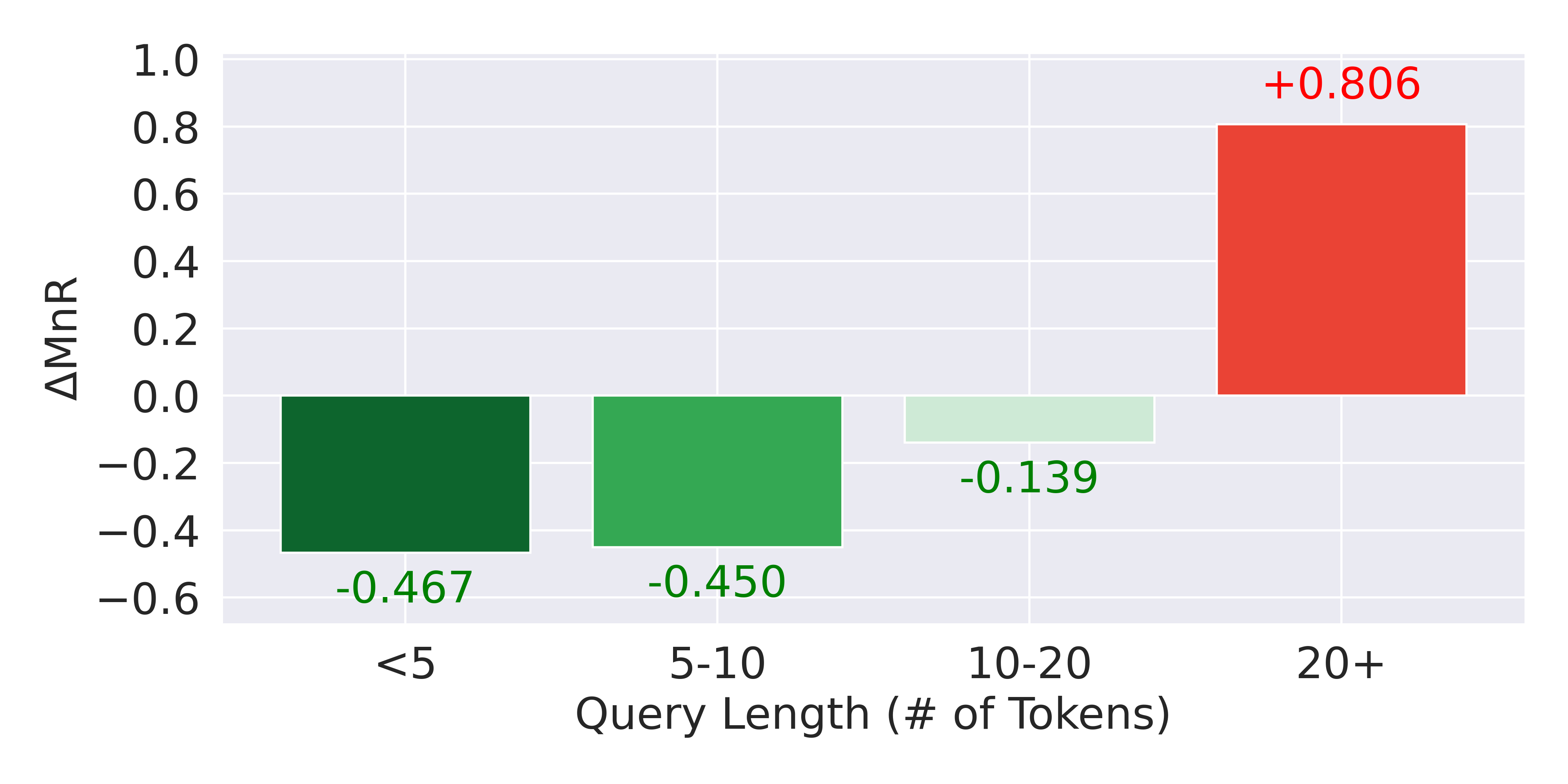}
    \vspace{-1em}
   \caption{Effect of soft query augmentation on MSR-VTT video ranks for queries of different lengths. The plot depicts average change in rank (lower is better) using a CLIP-B/32 backbone.}
   \label{fig:rank_change_query_length_barplot}
   \vspace{-0.5em}
\end{figure}

\begin{table}[t]
\begin{center}
\resizebox{0.95\linewidth}{!}{
\begin{tabular}{cccccc} 
\toprule
\textbf{Backbone} & \textbf{Query Aug.} & {R@1} & {R@5} & {R@10} & {nDCG} \\ 
\midrule
\multirow{2}{*}{CLIP-B/32}   
    & \xmark                           & 45.3     & 72.5    & 82.0  & 0.631 \\ 
    & \cmark                           & 48.1     & 74.9    & 83.9  & 0.652 \\
\midrule
\multirow{2}{*}{SigLIP-B/16} 
    & \xmark                           &  51.0 & 75.9  & 85.1   & 0.675  \\ 
    & \cmark                           & {51.5} & 76.3   & {85.5} & {0.677}  \\
\bottomrule
\end{tabular}}
\end{center}
\vspace{-6mm}
\caption{Effect of including pad tokens for soft query augmentation in MMS token-wise interaction. Results on MSR-VTT.}
\vspace{-1.0em}
\label{tab:query_padding}
\end{table}
\begin{table}[ht!]
\centering
\resizebox{\linewidth}{!}{
\begin{tabular}{lccc} 
\toprule
\textbf{Query} & \textbf{Before} $\downarrow$ & \textbf{After} $\downarrow$ & $\Delta$ $\downarrow$ \\
\midrule

a lady talks into a megaphone & 100 & 5 & \cellcolor{green!50}$-95$ \\
anchor talking about a shows & 97 & 46 & \cellcolor{green!40}$-51$ \\ 
a woman is talking about movies & 50 & 12 & \cellcolor{green!30}$-38$ \\
a man is dodging bombs & 100 & 66 & \cellcolor{green!20}$-34$ \\

a kid unwrapping his presents & 2 & 1 & \cellcolor{green!5}$-1$ \\
fox newscasters discuss chris christie and his poll numbers &  3 & 3 & \cellcolor{gray!10}$0$ \\
three woman doing a fashion show to music & 1 & 2 & \cellcolor{red!5} $+1$ \\

fast moving time is shown here & 2 & 54 & \cellcolor{red!20} $+52$ \\
a person is explaining something & 38 & 100 & \cellcolor{red!30} $+62$ \\
explainin about the scene in the net & 15 & 84 & \cellcolor{red!40}$+69$ \\
\bottomrule

\end{tabular}
}

\vspace{-0.5em}
\caption{The most improved and degraded queries after using soft query augmentation (for queries $<$20 tokens in length). 
\textbf{Before:} rank of the query's target video with no query augmentation. \textbf{After:} rank of the target video when performing query augmentation. 
$\Delta$: change in video rank after applying query augmentation. 
For all metrics, lower is better. The maximum rank is capped at 100.
}
\label{tab:query_expansion_qual}
\vspace{-1.5em}
\end{table}

In \cref{fig:rank_change_query_length_barplot} we perform a more in-depth analysis of what types of queries benefit most from soft augmentation. We observe a negative correlation between the query length and the use of expansions. Short queries of less than 20 tokens see the most benefit (as indicated by the mean rank of the target video), while queries longer than 20 tokens seem to be negatively impacted by query augmentation. One possible explanation for this phenomenon is that shorter, more abstract queries leave room for possible expansions while longer queries are either more descriptive or noisy. 

In \cref{tab:query_expansion_qual}, we highlight some of the most improved and most degraded queries as a result of query augmentation. We observe that the queries that benefit most seem to be more intuitive to conceptualize. For example a justifiable inference about an ``anchor talking about a shows'' might be the location of a news room or a news logo in the video frames. Queries with no or little change in their performance seem to be sufficiently descriptive to begin with. These queries target key features of the videos that can be matched with a variety of content without the need for expansion, and often perform well even without the expansions. However, we observe that queries with the most negative change from query expansions have the opposite properties to those that experience the largest gain. With query content like ``fast moving time'' and ``a person is explaining something,'' it is less obvious what additional search terms could be introduced to enhance retrieval of the target video.

\vspace{-1.5em}
\paragraph{Number of Sampled Frames.}
\begin{table}[]
    \centering
    \resizebox{0.8\linewidth}{!}{
    \begin{tabular}{ccccc}
    \toprule
         \textbf{\# of Frames} & {R@1} & {R@5} & {R@10} & {nDCG}\\
         \midrule 
         4 & 45.2 & 70.7 & 80.7 & 0.622 \\
         12 & 48.1 & 74.9 & 83.9 & 0.652 \\
         20 & 48.4 & 74.8 & 83.4 & 0.652 \\
    \bottomrule
    \end{tabular}}
    \vspace{-0.5em}
    \caption{Effect of number of sampled video frames on MSR-VTT retrieval performance using CLIP-B/32 backbone.}
    \label{tab:num_frames}
    \vspace{-0.5em}
\end{table}

In \cref{tab:num_frames} we assess the impact of the number of sampled video frames on retrieval performance. On the MSR-VTT dataset, we see only marginal improvements from increasing the number of frames beyond 12. However, it is important to keep in mind that this analysis is highly dataset-dependent. Retrieval tasks that depend on fine-grained motion will likely benefit from a higher sampling rate, while more spatially-heavy ones will see little improvements from denser frame selection.

\vspace{-1.2em}
\paragraph{Temporal Transformer Depth.}
\begin{table}[]
    \centering
    \resizebox{0.990\linewidth}{!}{
    \begin{tabular}{ccccc}
    \toprule
         \textbf{Temporal Transformer Depth} & {R@1} & {R@5} & {R@10} & {nDCG}\\
         \midrule 
         2 & 47.0 & 74.0 & 82.3 & 0.642 \\
         4 & 48.1 & 74.9 & 83.9 & 0.652 \\
         8 & 48.5 & 74.1 & 82.6 & 0.650  \\
    \bottomrule
    \end{tabular}}
    \vspace{-0.5em}
    \caption{Effect of \# of temporal transformer layers on MSR-VTT retrieval performance using CLIP-B/32 backbone.}
    \label{tab:tt_depth}
\vspace{-1em}
\end{table}

In \cref{tab:tt_depth} we experiment with different temporal transformer depths for forming the video features used in our tokenwise $\text{MMS}_V$ interaction. We find there to be a modest benefit from adding additional layers, suggesting that more powerful temporal modeling may be beneficial for retrieval. This finding, again, depends heavily on the types of videos being retrieved.

\section{Conclusions}
\label{sec:conclusions}
\vspace{-0.5em}
In this work, we introduced \methodname, a novel approach for text-to-video retrieval that uses efficient fine-grained interactions with both spatial and spatio-temporal visual features. Additionally,  \methodname is the first method to employ a sigmoid-based loss in T2VR, with our dual sigmoid loss formulation. We find that this interaction and training paradigm leads to strong representations for encoding spatial and temporal information while still being compatible when combined during retrieval. In the augmentations of our tokenwise interaction, we find that augmenting the query with padding tokens is beneficial for short queries, improving on the retrieval performance from pure query-to-video interaction. Additionally, we find that using the reciprocal rank fusion, an effective ranked retrieval fusion method, hurts retrieval performance, highlighting the potential for future exploration of deeper alignment between our scoring functions. 

\vspace{0.5em}
\noindent \small{{\textbf{Acknowledgments.} This research was partially supported by NSF GRF Grant No. DGE2139757 and by the Army Research Laboratory under Cooperative Agreement W911NF-21-2-0211. 
The views and conclusions contained in this document are those of the authors and should not be interpreted as representing the official policies, either expressed or implied, of the NSF, the Army Research Office or the U.S. Government. 
The U.S. Government is authorized to reproduce and distribute reprints for Government purposes notwithstanding any copyright notation herein.}}

{
    \small
    \bibliographystyle{ieeenat_fullname}
    \bibliography{zotero, main}
}

\clearpage
\setcounter{page}{1}
\maketitlesupplementary

\section{Additional Training Details}
\normalsize
We provide additional details about our training configurations in \cref{tab:supp_training_deets}. We also indicate dataset-specific settings, like batch size and number of epochs, in \cref{tab:supp_dataset_specific_deets}.

\begin{table}[ht!]
\centering
\resizebox{0.95\linewidth}{!}{\begin{tabular}{c|c}
\toprule
\textbf{Setting}        & \textbf{Value}            \\ \midrule

Learning Rate Schedule & Linear \\
Warmup Proportion (Linear) & 10\% \\
CLIP Param. Learning Rate & 1e-7 \\
Temporal Layer Learning Rate & 1e-4 \\
Optimizer & Adam  \\
Adam Betas & $\beta_1=0.9$, $\beta_2=0.98$ \\
Adam $\epsilon$ & 1e-6 \\
Weight Decay & 0.01 \\
Max. Grad. Norm & 1 \\
\bottomrule
\end{tabular}}
\vspace{-2mm}
\caption{Training settings for \methodname.}
\label{tab:supp_training_deets}
\end{table}

\begin{table}[ht!]
\begin{center}
\resizebox{0.95\linewidth}{!}{
\begin{tabular}{cccccc} 
\toprule
\textbf{Dataset} & \textbf{Backbone Type} & \textbf{Batch Size} & \textbf{Epochs} \\ 
\midrule
\multirow{2}{*}{MSR-VTT}   
    & ViT-B/32                          & 256     & 5     \\ 
    & ViT-B/16                         & 128     & 5    \\
\midrule
\multirow{2}{*}{MSVD}   
    & ViT-B/32                           & 256     & 5    \\ 
    & ViT-B/16                           & 128     & 5    \\
\midrule
\multirow{2}{*}{VATEX} 
    & ViT-B/32                          &  256 & 10    \\ 
    & ViT-B/16                           & 128 & 10  \\
\midrule
\multirow{2}{*}{DiDeMo} 
    & ViT-B/32                          &  64 & 20    \\ 
    & ViT-B/16                           & 64 & 20  \\
\midrule
\multirow{2}{*}{ActivityNet} 
    & ViT-B/32                          & 64  & 20    \\ 
    & ViT-B/16                           & 64 & 20  \\
\bottomrule
\end{tabular}}
\end{center}
\vspace{-6mm}
\caption{Dataset-specific training settings.}
\vspace{-2mm}
\label{tab:supp_dataset_specific_deets}
\end{table}

\section{Computational Analysis}

In \cref{tab:computation}, we analyze the computational trade-offs involved in the different MMS variants. We report numbers pertaining to both offline index creation and query-time ranking. During video indexing, we see that $\text{MMS}_{FV}$ is no more expensive than $\text{MMS}_V$, as they involve identical forward passes through the video encoder. At query-time, despite $\text{MMS}_{FV}$ involving more dot products, latency is virtually the same as the single-level interactions. Because query latency is dominated by the text encoding process (which involves self-attention), any differences in interaction complexity are rendered negligible. The main drawback of the two-level interaction is the storage cost of maintaining both spatial and spatiotemporal features in the video index, which can be mitigated by employing index compression methods.

\begin{table}[ht!]
\begin{center}
\resizebox{0.99\linewidth}{!}{
\begin{tabular}{l|ccc}
\toprule
  & \textbf{Indexing (ms/vid)} & \textbf{Query Latency (ms)} & \textbf{R@1} \\
\midrule
$\text{MMS}_F$      & 8.90 & 11.1 & 44.3 \\
$\text{MMS}_V$      & 9.64 & 11.1 & 47.0 \\
$\text{MMS}_{FV}$   & 9.64 & 11.2 & 48.1 \\
\bottomrule
\end{tabular}}
\end{center}
\vspace{-6mm}
\caption{Indexing time, query latency and retrieval accuracy on MSR-VTT 1K with CLIP-B/32 on A5000 GPU.}
\label{tab:computation}
\end{table}

During training, we find that the multi-level loss adds no additional computational burden. Backpropagation on our dual sigmoid loss involves the exact same number of gradient computations as doing so on a single-level loss (on $v$), and uses virtually the same amount of VRAM.

\section{Effect of Query Pad Token Choice}

In \cref{tab:supp_query_pad_tokens}, we show how video retrieval results are affected by different choices of padding token when using soft query augmentation in \methodname with a CLIP-B/32 backbone. Ordinarily (\eg when using only the special aggregation token to represent the query), the choice of padding token does not have any influence on retrieval outcomes. However, when performing soft query augmentation, all self-attention operations involve padding tokens, and the outputs of these extra tokens are used for interaction with visual features. As a result, the choice of pad token \textit{does} have an impact on retrieval results when using query augmentation and token-wise interaction. Because we freeze the token embeddings in the text encoder, we find that the choice of padding token has a noticeable effect on retrieval metrics. This is due to the fact that certain tokens will have pre-existing semantics that are better aligned with the query augmentation task than others. We found that the exclamation mark leads to the best performance out of the options we considered.

\begin{table}[ht!]
\begin{center}
\resizebox{0.999\linewidth}{!}{
\begin{tabular}{cccccc} 
\toprule
\textbf{Token ID} & \textbf{Token Text} & {R@1} & {R@5} & {R@10} & {nDCG} \\ 
\midrule
    31 & \code{@}                         & 46.0     & 74.6    & 83.3  & 0.644  \\ 
    49407 & \code{<|endoftext|>}          & 46.0     & 73.3    & 82.3  & 0.638  \\
    3002 & \code{...}                     &  47.8    & 74.6    & 83.6  & 0.652  \\ 
    13530 & \code{\_\_\_</w>}             & 47.9     & 72.8    & 83.5  & 0.646  \\
    49406 & \code{<|startoftext|>}        & 48.0     & 74.3    & \textbf{84.0}  & \textbf{0.653}  \\
    0 & \code{!}                          & \textbf{48.1} & \textbf{74.9}  & {83.9} & {0.652}     \\
\bottomrule
\end{tabular}}
\end{center}
\vspace{-6mm}
\caption{Effect of choice of padding token for soft query augmentation. Results on MSR-VTT using CLIP-B/32 backbone.}
\vspace{-2mm}
\label{tab:supp_query_pad_tokens}
\end{table}

\begin{figure*}[h!]
    \centering
    \includegraphics[width=\linewidth]{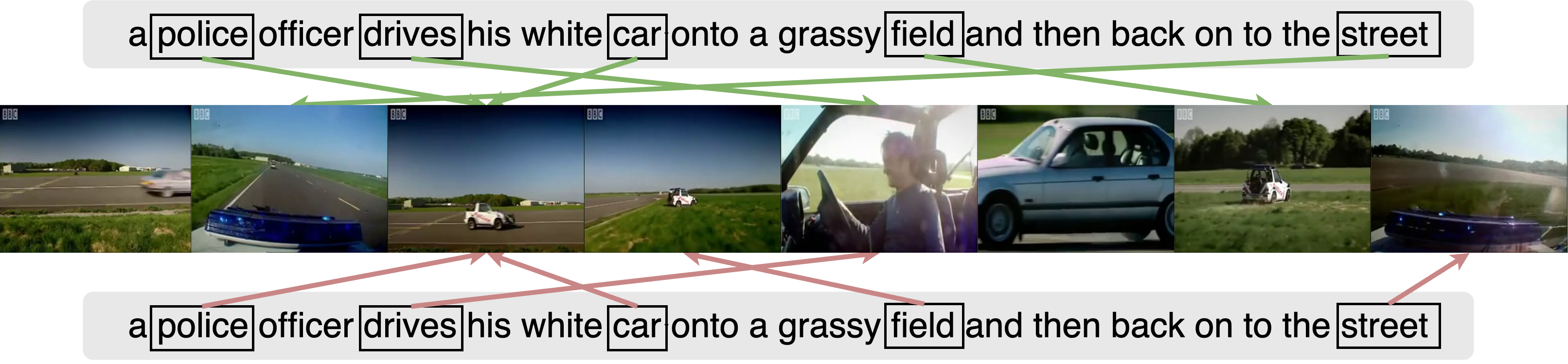}
    \caption{Visualization of the interactions between query tokens and video frames before and after the temporal encoder of \methodname, trained on MSR-VTT. The green arrow (\lineimg{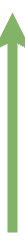}) represents the interaction between query tokens and frames \textbf{before} temporal encoding. The red arrow (\lineimg{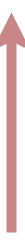}) represents the interaction between query tokens and frames \textbf{after} the temporal encoding.}
    \label{fig:visual}
\end{figure*}

\section{Visualization}
In \cref{fig:visual}, we explore how interactions between text tokens and frame representations change before and after the temporal transformer layers by visualizing the maximally similar frame to certain query tokens. To enhance the interpretability of this exploration, we do not use query or visual expansion during encoding. 
Generally, we find that the frame representations before and after the temporal encoder behave differently during interaction with the text tokens. In \cref{fig:visual}, the most obvious shift is in the similarities of ``field" and ``street." Prior to the temporal encoding, ``street" and ``field" correspond to frames that clearly represent the singular visual concept: a large grassy field with the car in the distance, and the street from the first person view of the car. After the temporal encoder, they then become associated with new frames: one with the car slightly on the grass field and another when the car is driving back onto the street. We interpret these results as a sign of stronger temporal contextualization in the frame representations after the encoding. 
Specifically, the associated frames seem to shift from depicting static concepts to more dynamic ones when temporally contextualized features are used.

\end{document}